\pdfoutput=1

\documentclass[11pt]{article}

\usepackage{acl}

\usepackage{times}
\usepackage{latexsym}
\usepackage{tikzsymbols}

\usepackage[T1]{fontenc}

\usepackage[utf8]{inputenc}

\usepackage{microtype}
\usepackage{pict2e}
\usepackage{amssymb}
\usepackage{amsmath}
\usepackage{graphicx}
\usepackage{booktabs}
\usepackage{multirow}
\usepackage{tabularx}
\def\utriang{\mbox{
\begin{picture}(5,7) 
\polyline(5,0)(5,5)(0,5)(5,0)
\end{picture}
}}

\def\ltriang{\mbox{
\begin{picture}(5,7)
\polyline(5,0)(0,0)(0,5)(5,0)
\end{picture}
}}

\newcommand{\look}{\includegraphics[height=1.4\fontcharht\font`X]{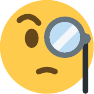}}
\newcommand{\ahead}{\includegraphics[height=1.4\fontcharht\font`X]{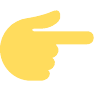}}
\newcommand{\memory}{\includegraphics[height=1.4\fontcharht\font`X]{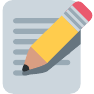}}
\title{LaMemo: Language Modeling with Look-Ahead Memory}

\author{Haozhe Ji$^\text{\look}$, Rongsheng Zhang$^\text{\ahead}$, Zhenyu Yang$^\text{\memory}$, Zhipeng Hu$^\text{\ahead}$, Minlie Huang$^\text{\look}$\thanks{\quad Corresponding author}  \\
$^\text{\look}${\small The CoAI group, DCST,
Institute for Artificial Intelligence, 
State Key Lab of Intelligent Technology and Systems, } \\ 
{\small Beijing National Research Center for Information Science and Technology, Tsinghua University, Beijing 100084, China}\\
$^\text{\ahead}${\small  Fuxi AI Lab, NetEase Inc., China}, $^\text{\memory}${\small OPPO Mobile Telecommunications Corp., Ltd, China }\\
  {\tt\small jhz20@mails.tsinghua.edu.cn,}
  {\tt\small aihuang@tsinghua.edu.cn } \\
  {\tt\small \{zhangrongsheng, zphu\}@corp.netease.com, }
  {\tt\small yangzhenyu@oppo.com} }

\begin{document}
\maketitle
\begin{abstract}
Although Transformers with fully connected self-attentions are powerful to model long-term dependencies, they are struggling to scale to long texts with thousands of words in language modeling. One of the solutions is to equip the model with a recurrence memory. However, existing approaches directly reuse hidden states from the previous segment that encodes contexts in a uni-directional way. As a result, this prohibits the memory to dynamically interact with the current context that provides up-to-date information for token prediction. To remedy this issue, we propose \textit{Look-Ahead Memory} ({LaMemo})\footnote{We are also inspired by the French word ``La Mémoire'', meaning ``the memory''.} that enhances the recurrence memory by incrementally attending to the right-side tokens, and interpolating with the old memory states to maintain long-term information in the history. 
LaMemo embraces bi-directional attention and segment recurrence with an additional computation overhead only linearly proportional to the memory length. Experiments on widely used language modeling benchmarks demonstrate its superiority over the baselines equipped with different types of memory.\footnote{Source code available at \url{https://github.com/thu-coai/LaMemo}.}
\end{abstract}

\section{Introduction}

Language modeling is an important task that tests the ability of modeling long-term dependencies by predicting the current token based on the previous context~\citep{penntreebank,wikitext103}. Recently, Transformer-based language models achieved remarkable performance by enabling direct interaction between long-distance word pairs. However, as the computation overhead grows with the length of the input sequence, 
Transformers can only process a fixed length segment at a time.
To allow long-term information flow across individual segments, existing approaches augment the model with a recurrence memory that stores hidden states computed in previous time steps~\citep{trm-xl} and their compressions~\citep{compressive,informer} for the target tokens to attend to. 

\begin{figure}[t!]
    \centering
    \includegraphics[width=0.9\columnwidth]{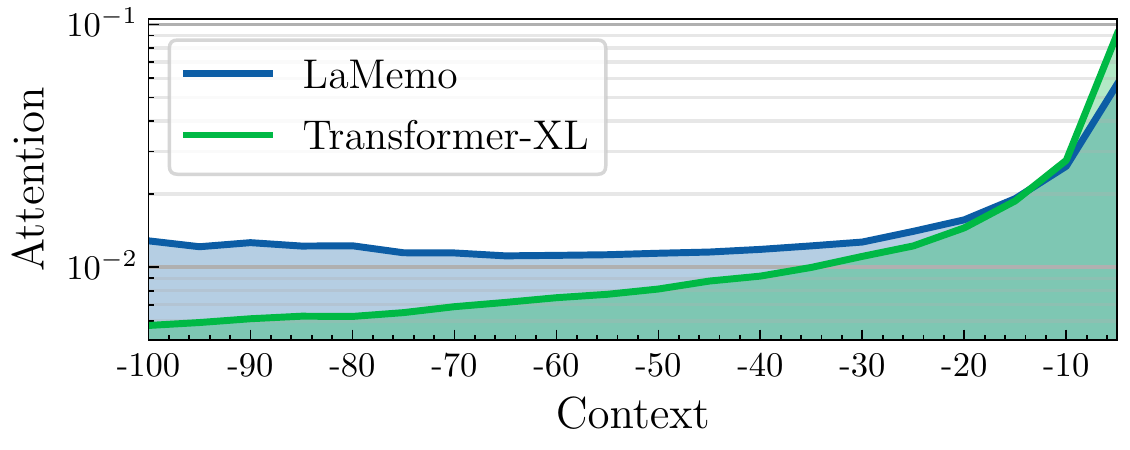}
    \caption{Attention weights on the context (in log-scale) in the final layer of Transformer-XL and LaMemo averaged on 15K tokens. Transformer-XL quickly loses attention to older contexts, while LaMemo maintains awareness to the history with the grow of the context length.}
    \label{fig:attn_utils}
\end{figure}

One limitation of this approach is that the recurrence memory is only aware of older contexts since they are previously computed to predict the next word from left to right. As a result, distant memory states become outdated and less activated by the current context, as illustrated in Figure \ref{fig:attn_utils}. 
When humans read or write a document, they maintain a memory that records important information from the past and often \textit{refresh} them under the current context to keep it up-to-date. 

In this paper, we propose \textit{Look-Ahead Memory} (LaMemo) where memory states ``look ahead'' to future time steps by attending to the token representations on their right side to provide up-to-date contextualization.\footnote{Note that the look-ahead attention does not exceed the current step of the autoregressive model to prevent information leakage.} 
To maintain information from the long-term history, we propose \textit{memory interpolation} to take both past and future tokens into consideration, which mimics the bi-directional attention.
Note that, directly applying bi-directional attention to update the memory representations brings an additional complexity of $\mathcal{O}(M^2)$ ($M$ is the memory length). This is expensive when the memory is very long. LaMemo incrementally attends to the right and accumulate the weighted attention sum from previous segments to simulate the full attention in only $\mathcal{O}(M\times N)$ complexity ($N$ is the target sequence length), which does not increase the attention complexity of Transformer-XL, namely $\mathcal{O}(N^2+M\times N)$. We provide an illustration of this mechanism in Figure \ref{fig:lahm_mech}.

Another technique proved to be effective in language modeling is the relative positional encoding~\citep{shaw_rpe, DBLP:journals/corr/abs-1809-04281, trm-xl}, which biases the pair-wise attention score purely based on the relative distance of the two tokens. However its ability to generalize to the attention of the future tokens remains unknown, since both the distance and the direction need to be taken into consideration. In preliminary experiments, we observed the unstability of directly applying the relative positional encoding of \citet{trm-xl} to this setting. 
We propose a simple yet effective modification based on \citet{trm-xl} that disentangles the bias of the relative distance and the attention direction which facilitates the training of LaMemo.
We give both theoretical and empirical analysis to the unstability issue and demonstrate the effectiveness of the proposed disentangled relative positional encoding method.

To sum up, our contributions are as follows:

(1) We propose LaMemo, a memory mechanism that 
incrementally attends to the right-side tokens, 
and interpolates with the old memory, which enables \textit{bi-directional} interaction 
with a complexity linear in memory length.

(2) We propose disentangled relative positional encoding, a simple yet effective solution that disentangles the relative distance and the attention direction that can better generalize to the attention of the future tokens. 

(3) We conduct experiments on standard language modeling benchmarks and demonstrate LaMemo's superiority over various baselines equppied with different types of memory mechanisms, despite some having an access to longer contexts. Comprehensive comparisons show the benefits of learning memory representations contextualized with up-to-date information.

\section{Background}

\subsection{Transformer for Language Modeling}

A Transformer~\citep{vaswani} is composed of multiple layers of identical blocks, including a multi-head self-attention~\citep{bahdanau} that calculates pair-wise token interaction and a feed-foward layer for position-wise projection with a non-linear activation. Both two modules are followed by residual connections~\citep{resnet} and layer normalization~\citep{layernorm} to facilitate optimization. 

Given the input sequence representations of the current $\tau$-th segment $\boldsymbol{X}_\tau=[\boldsymbol{x}_{\tau+1},\cdots,\boldsymbol{x}_{\tau+N}]\in \mathbb{R}^{N\times d}$ where $N$ is the target sequence length and $d$ is the hidden state size, they are first mapped into queries $Q$, keys $K$ and values $V$ by learned weight matrix to compute self-attention:
\begin{equation}\label{eq:qkv}
    \boldsymbol{Q}_\tau=\boldsymbol{X}_\tau \boldsymbol{W}_q, \boldsymbol{K}_\tau=\boldsymbol{X}_\tau \boldsymbol{W}_k, \boldsymbol{V}_\tau=\boldsymbol{X}_\tau \boldsymbol{W}_v,
\end{equation}
where $\boldsymbol{W}_q,\boldsymbol{W}_k,\boldsymbol{W}_v\in \mathbb{R}^{d\times d}$ are learnable projection matrices. To perform multi-head self-attention, $Q, K, V$ are further split into $H$ heads. For simplicity, we only consider the case of a single head.
In language modeling, the attention map is always added by a causal mask to avoid information leakage from the future when predicting the next token:
\begin{align}\label{equ:causal}
    \boldsymbol{C}^{\rightarrow}_\tau&=\text{Causal-Attn}(\boldsymbol{Q}_\tau,\boldsymbol{K}_\tau,\boldsymbol{V}_\tau)\nonumber\\
    &=\text{softmax}_{\ltriang}\Big(\frac{\boldsymbol{Q}_\tau \boldsymbol{K}_\tau^\top}{\sqrt{d}}\Big)\boldsymbol{V}_\tau,
\end{align}
where $\text{softmax}_{\ltriang}(\cdot)$ masks position $j>i$ for the $i$-th row of the input matrix with $-\infty$ before taking the softmax. The resulted context representations are concatenated and then projected to the final outputs $\boldsymbol{O}_\tau\in\mathbb{R}^{N\times d}$ with a learnable projection matrix $\boldsymbol{W}_o\in \mathbb{R}^{d\times d}$.
Finally, the self-attention outputs $\boldsymbol{O}_\tau$ are added by the input representations $\boldsymbol{X}_\tau$ and fed to the following point-wise non-linear transformation, denoted as $f(\cdot)$:
\begin{align}\label{equ:non-linear}
    f(\boldsymbol{x}) = \textrm{LN}\Big(\text{FFN}\big(\text{LN}(\boldsymbol{x})\big) + \text{LN}(\boldsymbol{x})\Big),
\end{align}
where $\text{LN}(\cdot)$ is the layer normalization and $\text{FFN}(\cdot)$ is the feed-forward layer, both of which are applied to each row vector individually. The final output of this Transformer layer is $f(\boldsymbol{O}_\tau + \boldsymbol{X}_\tau)$.

Outputs of the final layer are projected to the vocabulary to predict $\Pr(w_t|w_1,\cdots,w_{t-1})$. The joint probability of predicting the whole segment is the product of these conditional factors. The final objective is to maximize the following log-likelihood:
\begin{equation}
    \log \Pr(\boldsymbol{w})=\prod_{t=1}^{N} \log \Pr(w_t|w_1,\cdots, w_{t-1}).
\end{equation}

\begin{figure}
    \centering
    \includegraphics[width=0.5\columnwidth]{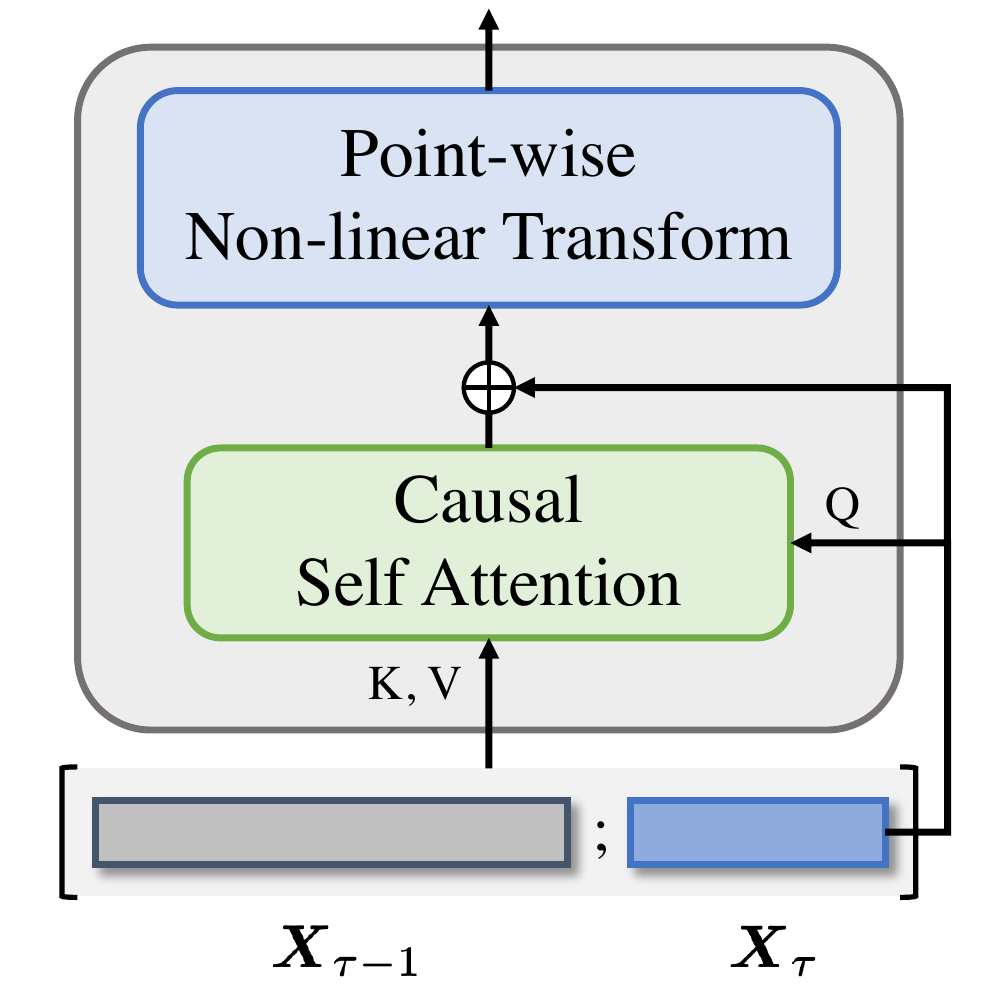}
    \caption{The architecture of Transformer-XL augmenting with a recurrence memory.}
    \label{fig:trm_xl_arch}
\end{figure}

\subsection{Recurrence Memory Mechanism}

To enable the Transformer to consider more contextual information from previous segments, \citet{trm-xl} proposed to augment the Transformer with a recurrence memory which stores the hidden states of previous time steps as extended keys and values, as shown in Figure \ref{fig:trm_xl_arch}. Concretely, let us consider a memory length of $M$ and memory representations $\boldsymbol{X}_{\tau-1}=[\boldsymbol{x}_{\tau-M+1},\cdots, \boldsymbol{x}_{\tau}]\in \mathbb{R}^{M\times d}$. The extended key and value matrices are obtained by prepend $\boldsymbol{X}_{\tau-1}$ to $\boldsymbol{X}_\tau$ before projection:
\begin{align}
\tilde{\boldsymbol{X}}^{\text{sg}}_\tau &= [\text{sg}(\boldsymbol{X}_{\tau-1})\circ \boldsymbol{X}_\tau]\in\mathbb{R}^{(M+N)\times d},
\end{align}
where $\text{sg}(\cdot)$ stands for stop-gradient which disables gradient propagation to previous segments, and $[\cdot \circ {\cdot}]$ indicates concatenation of hidden states along the length dimension. Extended by the recurrence memory, each query vector can consider contexts even beyond the total context length of the attention $M+N$. As illustrated by \citet{trm-xl}, the effective context length grows linearly to the number of layers and the attention context length due to layer-wise reusing.

Another technique necessary to the recurrence memory is the relative positional encodings. By considering only the relative distance between two tokens when computing the attention score, it avoids temporal confusion caused by indexing the same position across segments and injects useful relative bias. Transformer-XL uses the fixed sinusoidal encoding matrix~\citep{vaswani} to provide relative distance bias and learns global bias terms shared across different layers, which can extrapolate to longer contexts with a great reduction of parameters compared to \citet{shaw_rpe}:
\begin{align}\label{equ:xl-rpe}
    \boldsymbol{A}^{{xl}}_{i,j} &= \boldsymbol{X}_i^\top \boldsymbol{W}_q^\top \boldsymbol{W}_{k}^E \boldsymbol{X}_j + \boldsymbol{X}_i^\top \boldsymbol{W}_q^\top \boldsymbol{W}_{k}^R \boldsymbol{R}_{i-j} \nonumber \\
    &+ \boldsymbol{u}^\top \boldsymbol{W}_{k}^E \boldsymbol{X}_j + \boldsymbol{v}^\top \boldsymbol{W}_{k}^R \boldsymbol{R}_{i-j},
\end{align}
where $\boldsymbol{R}$ is the sinusoid encoding matrix, $\boldsymbol{u}, \boldsymbol{v}$ are learnable weight vectors governing the global content and position bias, and $\boldsymbol{W}_{k}^E, \boldsymbol{W}_{k}^R$ are separate key projection matrices for the content and position respectively.

\begin{figure}[t]
    \centering
    \includegraphics[width=0.9\columnwidth]{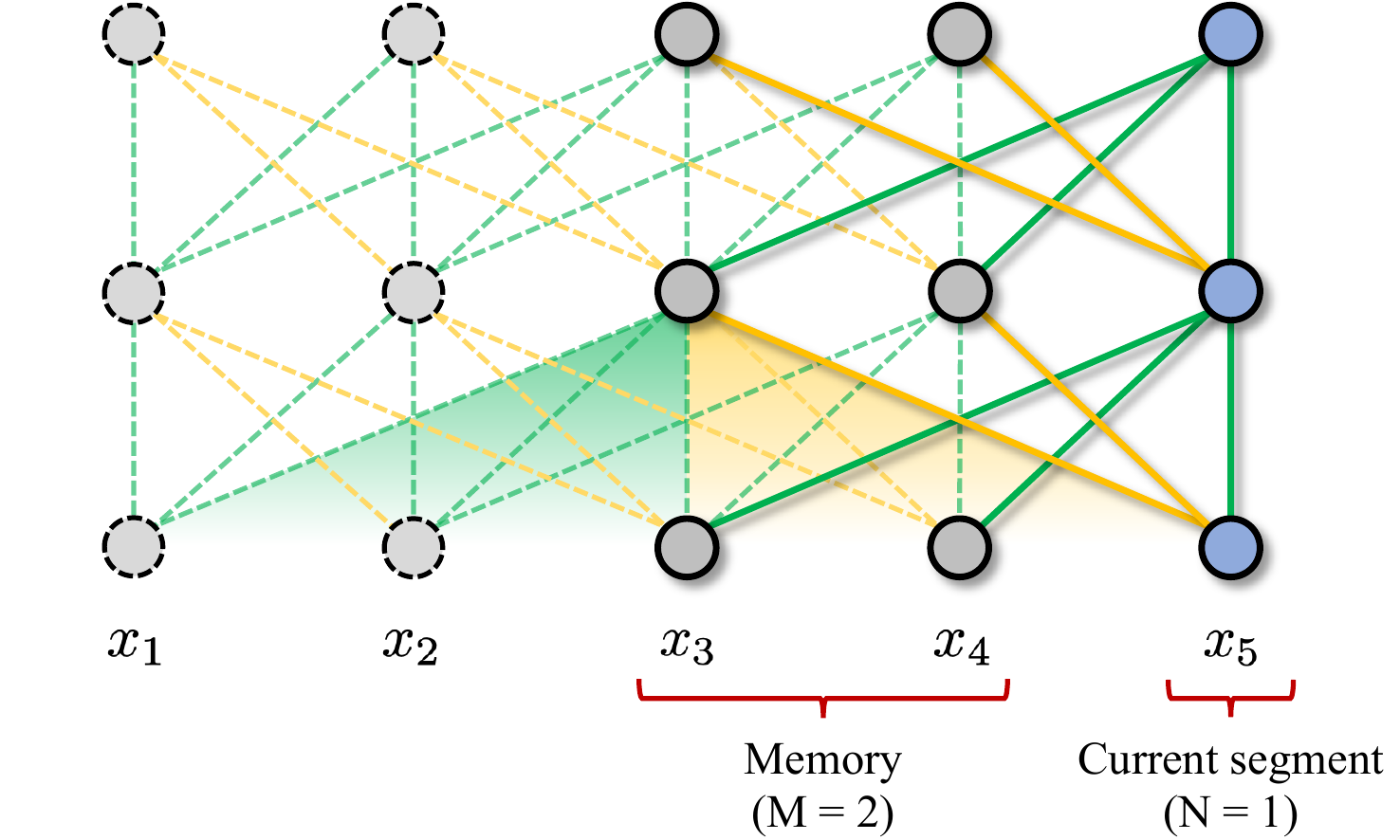}
    \caption{Illustration of LaMemo with a memory length $M=2$ and a target sequence length $N=1$ for clarity. Solid lines stand for the attention connections computed at this iteration while dashed lines represent the previously computed attention.}
    \label{fig:lahm_mech}
\end{figure}

\section{Method}

In this section, we describe our method in detail with our motivation to learn better representations for the memory.

\subsection{Look-Ahead Attention}\label{sec:lahm}

Human language is sequential with one word following another, but humans process information usually in a non-sequential way and re-contextualize certain contents for several times. For example, when countering complicated contents during reading, humans usually first store them temporarily in the memory and continue to scan for relevant information if any, and revisit those old contents to refresh their meaning quite often. This dynamic memory refreshing mechanism enables us to thoroughly understand the passage under current contexts. 

Existing recurrence memory however, lacks this dynamic contextualization ability. As the representations in the recurrence memory are previously computed conditioned on their past, they are not aware of the current contexts which provide more relevant information for the current token prediction.

To address this limitation, we propose a look-ahead attention that allow the memory to attend to the contexts on their right. Formally, we reuse the notation $\boldsymbol{X}_\tau=[\boldsymbol{x}_{\tau+1},\cdots,\boldsymbol{x}_{\tau+N}] \in \mathbb{R}^{N\times d}$ for the representations of the current target sequence and $\boldsymbol{X}_{\tau -1}=[\boldsymbol{x}_{\tau-M+1},\cdots,\boldsymbol{x}_\tau]\in \mathbb{R}^{M\times d}$ for the representations of the memory.

Let us consider the $i$-th position of the memory $\boldsymbol{X}_{\tau-1}$, $\boldsymbol{x}_i$ can attend to position $\boldsymbol{x}_j$ on its right ($j>i$) without causing information leakage as long as $j\le \tau+1$. 
Though appealing, this naïve approach requires to calculate an $M$ by $M$ attention map, which would become inefficient and redundant when $M$ is significantly greater than $N$. 
Actually, since the target segment moves forward $N$ positions at each iteration, we devise an incremental manner of look-ahead attention computation that only requires the newest $N$ positions on the right as key-value pairs.
\begin{equation}
    \tilde{\boldsymbol{X}}_{\tau-1} = [{\boldsymbol{x}}_{\tau-N+2},\cdots,{\boldsymbol{x}}_{\tau+1}]\in \mathbb{R}^{N\times d}.
\end{equation}
Then the look-ahead attention results computed previously can be effectively reused and interpolated with the current ones (\S{\ref{sec:mem_inter}}). Concretely, we formalize the look-ahead attention as follows:
\begin{align}
\tilde{\boldsymbol{K}}_{\tau-1}&=\tilde{\boldsymbol{X}}_{\tau-1} \boldsymbol{W}_k, \tilde{\boldsymbol{V}}_{\tau-1}=\tilde{\boldsymbol{X}}_{\tau-1} \boldsymbol{W}_v,\\
\label{equ:lahm}
\boldsymbol{C}^{\leftarrow}_{\tau-1}&=\text{LookAhead-Attn}(\boldsymbol{Q}_{\tau-1}, \tilde{\boldsymbol{K}}_{\tau-1}, \tilde{\boldsymbol{V}}_{\tau-1})\nonumber\\
&=\text{softmax}_{\utriang}\Big(\frac{\boldsymbol{Q}_{\tau-1}\tilde{\boldsymbol{K}}_{\tau-1}^\top}{\sqrt{d}}\Big)\tilde{\boldsymbol{V}}_{\tau-1},
\end{align}
where $\text{softmax}_{\utriang}(\cdot)$ masks position $j\le i$ for the $i$-th row of the input matrix with $-\infty$ before softmax. $\boldsymbol{Q}_{\tau-1}$ is obtained by Eq. (\ref{eq:qkv}), and the projection matrices of query, key and value are all shared with the causal attention. We illustrate this in Figure \ref{fig:lahm_mech} where the look-ahead attention (yello paths) increases the attention window of each memory state to $M$ tokens on its right.

\begin{figure}
    \centering
    \includegraphics[width=0.95\columnwidth]{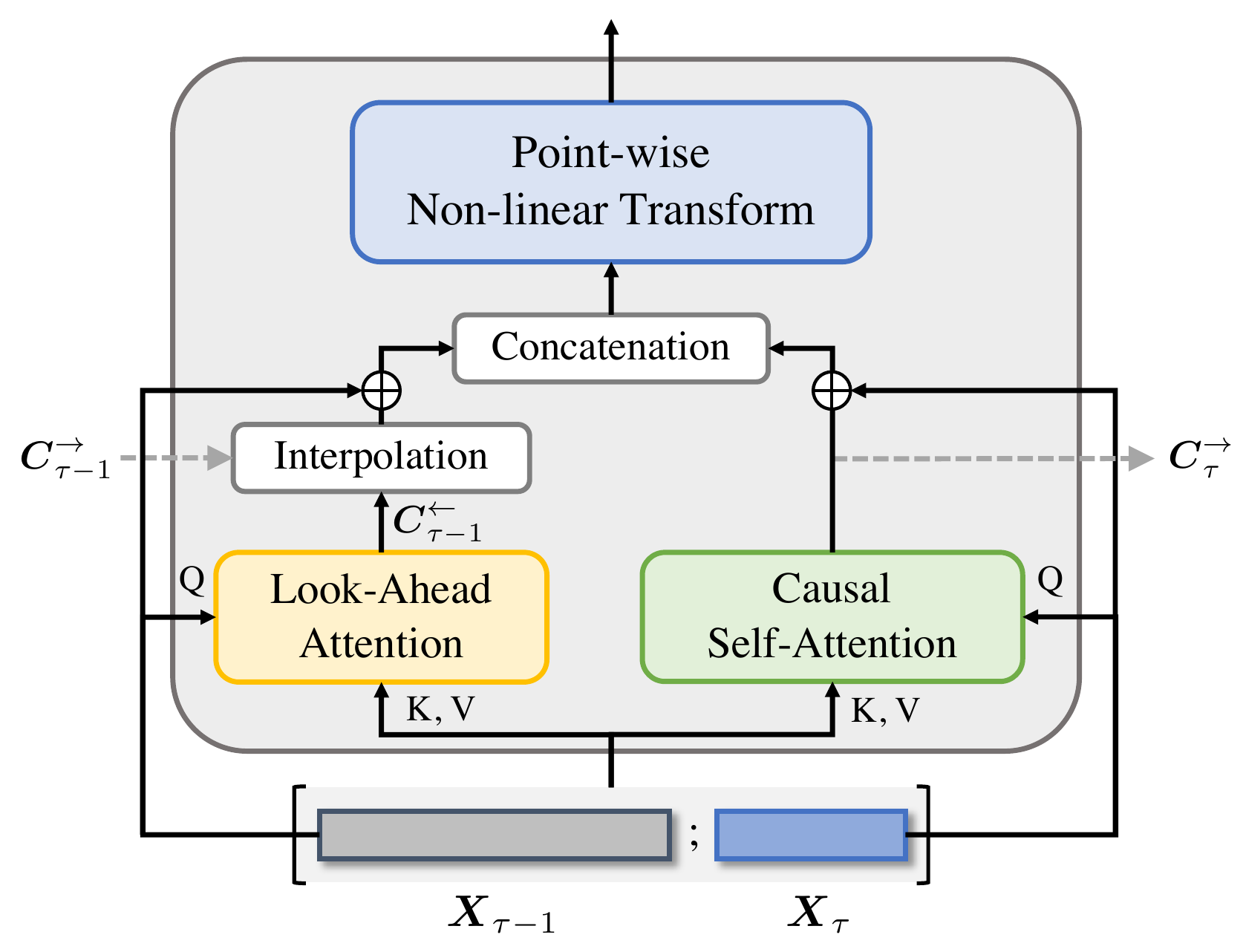}
    \caption{The architecture of LaMemo with look-ahead attention and memory interpolation that refresh the memory dynamically with both the current contexts and the long-term history.}
    \label{fig:lahm_arch}
\end{figure}

\subsection{Memory Interpolation}\label{sec:mem_inter}

To save computations for looking-ahead and effectively reuse the attention results of the past, we propose memory interpolation that smoothly interpolates attention results from both the future and the past to provide bi-directional contextualization.

Recall that in the previous iteration, we have calculated the causal context representations $\boldsymbol{C}^{\rightarrow}_{\tau-1}$ of $\boldsymbol{X}_{\tau-1}$ using Eq. \ref{equ:causal}, where each row is a linear combination of the weighted token representations of the previous tokens. In Sec. \ref{sec:lahm}, we describe the look-ahead attention which enables $\boldsymbol{X}_{\tau-1}$ to attend to the contexts on their right and computes $\boldsymbol{C}^{\leftarrow}_{\tau-1}$ using Eq. \ref{equ:lahm}. Here, we formulate the memory interpolation as the interpolation between the old representations $\boldsymbol{C}^{\rightarrow}_{\tau-1}$ and the new ones $\boldsymbol{C}^{\leftarrow}_{\tau-1}$ with a coefficient vector $\boldsymbol{\alpha}_{\tau-1}\in \mathbb{R}^{M}$ controlling the memorization of the past activations:
\begin{align}\label{equ:mem-interp}
    \boldsymbol{C}^{\leftrightarrow}_{\tau-1}&=\text{Mem-Interp}(\boldsymbol{C}^{\rightarrow}_{\tau-1}, \boldsymbol{C}^{\leftarrow}_{\tau-1}, \boldsymbol{\alpha}_{\tau-1})\nonumber\\
    &=\boldsymbol{\alpha}_{\tau-1}\text{sg}(\boldsymbol{C}^{\rightarrow}_{\tau-1}) + (1-\boldsymbol{\alpha}_{\tau-1})\boldsymbol{C}^{\leftarrow}_{\tau-1}.
\end{align}
The resulted $\boldsymbol{C}^{\leftrightarrow}_{\tau-1}$ which attend to contexts from both directions, are further fed to the non-linear transformation defined in Eq. \ref{equ:non-linear} to update representations in higher layers.

For $\boldsymbol{\alpha}_{\tau-1}$, we define it to be the sum of the normalized attention weights on the previous tokens when calculating $\boldsymbol{C}^{\rightarrow}_{\tau-1}$ (Eq. \ref{equ:causal}):
\begin{equation}\label{equ:alpha}
    \boldsymbol{\alpha}_{\tau-1}=\frac{\text{sg}(\boldsymbol{s}^{\rightarrow}_{\tau-1})}{\text{sg}(\boldsymbol{s}^{\rightarrow}_{\tau-1}) + \boldsymbol{s}^{\leftarrow}_{\tau-1} + \varepsilon},
\end{equation}
where $\boldsymbol{s}^{\rightarrow}_{\tau-1}$ is the sum of the unnormalized attention score of $\boldsymbol{C}^{\rightarrow}_{\tau-1}$, which is the denominator of the softmax in Eq. \ref{equ:causal}. Similarly, $\boldsymbol{s}^{\leftarrow}_{\tau-1}$ is the denominator of the softmax in Eq. \ref{equ:lahm}. $\varepsilon$ is a small value to prevent zero division error in practice. Then Eq. \ref{equ:mem-interp} can be derived into a form that resembles the bi-directional attention with the queries attending to positions on both sides\footnote{Note that the query vectors for the past and the future are under different contextualization in higher layers of the model.} (Appendix \ref{app:derive}). Figure \ref{fig:lahm_arch} shows the architecture of LaMemo.


Note that the difference between the hidden state reuse in the recurrence memory and our memory interpolation is that they simply reuse the static representations to extend the contexts for attention while we update the memory representations by aggregating weighted attention sum of the history without the need to recompute them.

\subsection{Disentangled Relative Positional Encodings}\label{sec:dis-rpe}

As the look-ahead attention allows the memory to attend to future tokens on its right, we need a relative positional encoding scheme that can generalize to this setting. We start by considering the relative positional encoding in Transformer-XL, as described by Eq. \ref{equ:xl-rpe}. When the $i$-th query vector attending to a position $j=i+\Delta > i$, we have $\boldsymbol{R}_{i-j}=\boldsymbol{R}_{-\Delta}$. As defined by \citet{vaswani}, $\boldsymbol{R}_\Delta\in\mathbb{R}^D$ is composed of sine and cosine functions with different frequencies. Since the sine function is odd, $\sin(-\omega\Delta)=-\sin(\omega\Delta)$, we have $\boldsymbol{R}_{-\Delta}\ne \boldsymbol{R}_{\Delta}$ 
so that it can represent attention in different directions ($\pm$ sign of $\Delta$) with the same relative distance (absolute value of $\Delta$). 

However, this approach \textit{solely} relies on the fixed sinusoid encodings to represent the relative distance and the attention direction. 
We argue that disentangling them is more effective in capturing these two types of temporal biases and also mitigates the numerical unstability issue. Specifically, we propose to learn two direction-aware global position biases to parameterize the \textit{sign} and query $\boldsymbol{R}$ with the \textit{absolute} value of the relative distance:
\begin{align}
    \boldsymbol{A}^{{dis}}_{i,j} &= \boldsymbol{X}_i^\top \boldsymbol{W}_q^\top \boldsymbol{W}_{k}^E \boldsymbol{X}_j + \boldsymbol{X}_i^\top \boldsymbol{W}_q^\top \boldsymbol{W}_{k}^R \boldsymbol{R}_{|i-j|} \nonumber \\
    &+ \boldsymbol{u}^\top \boldsymbol{W}_{k}^E \boldsymbol{X}_j + \boldsymbol{v}_{i-j}^\top \boldsymbol{W}_{k}^R \boldsymbol{R}_{|i-j|},
\end{align}
where $\boldsymbol{v}_{i-j} = \boldsymbol{v}_+$ if $i\ge j$ else $\boldsymbol{v}_-$. 
The global positional bias now explicitly separates the contributions of $\text{sgn}(i-j)$ and $|i-j|$, which can better generalize to long distance in both forward and backward directions. 

To illustrate the numerical unstability caused by adapting Eq. \ref{equ:xl-rpe} to $j>i$, we derive the variance of the dot product $\boldsymbol{x}^T\boldsymbol{R}_{i-j}$ where $\boldsymbol{x}$ is a random vector.
We show that the variance undergoes an oscillation and cannot be properly bounded everywhere when $i$ shifts from $i\ge j$ to $i<j$. Detailed analysis are presented in Appendix \ref{app:unstability}.


\section{Experiments}

We evaluate LaMemo on both word-level and character-level language modeling tasks and compare with existing Transformer baselines augmented with different types of memory.

\subsection{Datasets and Metrics}

For word-level language modeling task, we consider \textbf{Wikitext-103}~\citep{wikitext103}, which is the most widely used word-level language modeling benchmark. It contains 103 million tokens for training from 28 thousand wikipedia articles, with an average length of 3.6 thousand tokens per article and a vocabulary size around 260K. We report perplexity (ppl) on the dev and test set.

We also evaluate on two character-level language modeling benchmarks \textbf{enwik8} and \textbf{text8}~\citep{mahoney2011large}. Both datasets contain 100 million Wikipedia characters. While \textbf{enwik8} is unprocessed, \textbf{text8} is preprocessed by case lowering and filtering to include only 26 letters from \texttt{a} to \texttt{z} and space. On both datasets, we report bit per character (bpc) on the dev and test set. 

\subsection{Baselines}

To directly compare with different types of memory, we consider Transformer-XL and its variations with the same model architecture but different memory mechanism.

\textbf{Transformer+RPE} is the vanilla Transformer~\citep{vaswani} that uses relative positional encodings from \citet{trm-xl} but does not extend the context with additional memory.

\textbf{Transformer-XL}~\citep{trm-xl} is a Transformer model equipped with relative positional encodings and a recurrence memory comprised of hidden states computed in previous time steps to extend the context length of the attention.

\textbf{Compressive Transformer}~\citep{compressive} extends Transformer-XL with an external compressive memory that stores compressed hidden states at the temporal level using convolutional networks.

\textbf{$\infty$-former}~\citep{informer} 
uses continuous space attention to attend over the external memory which consists of continuous signals. They also updated the external memory with recent hidden states to enable unbounded memory capacity. 

\subsection{Implementation Details}

\begin{table*}[t!]
    \centering
    \small
    \begin{tabular}{l cccccc}
    \toprule[1pt]
        Model & \#Params & Mem size & Ext mem size & \#FLOPS & dev ppl & test ppl \\
        \midrule[0.5pt]
        Transformer+RPE& 151M & 0 & 0 & 148M & 28.11 & 29.14 \\
        Transformer-XL~\citep{trm-xl} & 151M & 150 & 0 & {157M} & {23.42} & 24.56 \\
        Compressive Transformer~\citep{compressive} & 161M & 150 & 150 & 169M & - & 24.41 \\
        $\infty$-former~\citep{informer} & 160M & 150 & 150 & 235M & - & 24.22 \\
        LaMemo & 151M & 150 & 0 & 191M & \textbf{22.98} & \textbf{23.77} \\
    \bottomrule[1pt]
    \end{tabular}
    \caption{Word-level language modeling results on Wikitext-103. We report ppl (\textit{perplexity}) on dev and test set. We also report the number of parameters, memory size, external memory size, and the number of FLOPS (\textit{floating-point operations}) for computing one step prediction on average.}
    \label{tab:wiki103}
\end{table*}

We follow the standard architecture of the Transformer-XL~\citep{trm-xl} that has different configurations for different tasks. Specifically, on Wikitext-103, we use a 16-layer Transformer with 10 attention heads and head dimension 41 equipped with adaptive embeddings~\citep{adaptive_input}. We control the target sequence length to be 150 and the memory length 150 for all models following the setting of \citet{trm-xl}. For the Compressive Transformer and $\infty$-former, we additionally use an external memory of size 150 following the setting of \citet{informer}.\footnote{The external memory consists of 150 compressed vectors for Compressive Transformer, and 150 radial basis functions for $\infty$-former respectively.} On the text8 and enwik8 datasets, we use a 12-layer Transformer with 8 heads and head dimension 64. The length of the target sequence and the recurrence memory are both set to 512. In the main results we use the identical evaluation setting to the training phase on all datasets and do not use a longer memory. 
We use the Pytorch framework~\citep{pytorch} and Apex for mixed-precision training. In practice, we found that calculating the exponentials (\S{\ref{sec:mem_inter}}) may lead to numerical overflow in mixed-precision mode, so we compute the logarithm of the exponential sum using \texttt{logsumexp} and \texttt{logaddexp} operator. Further details of the dataset and the hyperparameter settings are described in the Appendix \ref{app:experiment}.

\subsection{Main Results}

\begin{table}[t!]
    \centering
    \small
    \begin{tabular}{lcc}
    \toprule[1pt]
    Model & dev bpc & test bpc \\
    \midrule[0.5pt]
    \multicolumn{3}{c}{Dataset: \textit{text8}}\\
    \midrule[0.5pt]
    Transformer+RPE & 1.232 & 1.303 \\
    Transformer-XL~\citep{trm-xl} & 1.172 & 1.239 \\
    LaMemo & \textbf{1.128} & \textbf{1.196} \\
    \midrule[0.5pt]
    \multicolumn{3}{c}{Dataset: \textit{enwik8}}\\
    \midrule[0.5pt]
    Transformer+RPE & 1.253 & 1.240 \\
    Transformer-XL~\citep{trm-xl} & 1.150 & 1.128 \\
    LaMemo & \textbf{1.129} & \textbf{1.107} \\
    \bottomrule[1pt]
    \end{tabular}
    \caption{Character-level language modeling results on text8 and enwik8. We report bpc (\textit{bits-per-character}) on the dev and test set.}
    \label{tab:text8}
\end{table}

We show the results of word-level language modeling benchmark Wikitext-103 in Table \ref{tab:wiki103}. We first observe that all the models extended with memories significantly outperforms Transformer+RPE. Under the same memory length, LaMemo outperforms Transformer-XL with a clear margin, which demonstrates the effectiveness of learning dynamic memory representations over static ones. When compared to the compressive memory and the unbounded memory that take longer contexts into account, LaMemo still achieves lower perplexity. This indicates that the look-ahead memory allows the language model to exploit the recent contexts to gain performance, while simply increasing the context length yields marginal improvement. 
This is in accordance with previous findings of how language models utilize contexts~\citep{sharp_near,Do-lr}. In terms of the parameters, LaMemo has the same number of parameters as the Transformer-XL while other baselines use additional parameters in CNN to compress or smooth the hidden states. Lastly, we show the number of FLOPS necessary for computing one step prediction. $\infty$-former has the highest number of FLOPS for resampling enough points from the continuous signal to update the memory using smoothing techniques. LaMemo also incurs additional computations to re-contextualize the memory under the current context. Note that although the Compressive Transformer has lower number of FLOPS than LaMemo, it has an external memory that consumes more GPU memory.

We also present the results of character-level language modeling on text8 and enwik8 datasets in Table \ref{tab:text8}. We observe similar trends as the results on the word-level benchmark, where LaMemo outperforms Transformer-XL by 0.04 on text8 and 0.02 on enwik8 with the same context length. Additionally, we observe that all models exhibit overfitting on text8, which might be caused by the extremely small vocabulary size of the dataset.

\subsection{Ablation Study}

We conduct ablation studies on Wikitext-103 to examine the effects of the proposed techniques, i.e., look-ahead attention, memory interpolation, and disentangled relative positional encodings.

We use the same model achitecture and the same target and memory length as the main results. We first study three configurations, including (1) using the \textbf{Full} model setting, (2) ablating the memory interpolation module (\textbf{w/o mem interp}), i.e., set the memorizing coeffecient $\boldsymbol{\alpha}_{\tau-1}=0$, and (3) ablating the look-ahead attention (\textbf{w/o look-ahead}), i.e., only use the causal context representations $\boldsymbol{C}^{\rightarrow}_{\tau-1}$ in each layer. As shown in the First three rows in Table \ref{tab:ablation}, both the memory interpolation and the look-ahead attention are indispensible for achieving the best performance. Additionaly, we found that cancelling out memory interpolation 
leads to a worse performance, which indicates that the distant past still provides additional information beyond the current context. 

\begin{table}[t!]
    \centering
    \small
    \begin{tabular}{lccc}
    \toprule[1pt]
    Configuration & Encoding & dev ppl & test ppl \\
    \midrule[0.5pt]
    Full & Ours & 22.98 & 23.77 \\
    w/o mem interp & Ours & 23.67 & 24.90 \\
    w/o look-ahead & Ours & 23.42 & 24.56 \\
    Full & \citet{trm-xl} & FAIL & FAIL \\
    \bottomrule[1pt]
    \end{tabular}
    \caption{Ablation study on Wikitext-103. We investigate three model configurations and two encoding schemes.}
    \label{tab:ablation}
\end{table}

The second study targets at studying different encoding schemes. We substitute our encodings with the RPE of Transformer-XL~\citet{trm-xl} and run multiple experiments with 3 different random seeds, but all the models fail to converge. 
We plot the training curves using two encodings in Figure \ref{fig:loss-comp} in Appendix \ref{app:unstability}, where we observe that our disentangled RPE is more stable during training and achieves lower perplexity.

\section{Extrapolating to Longer Contexts}

\begin{figure}[t!]
    \centering
    \includegraphics[width=0.8\columnwidth]{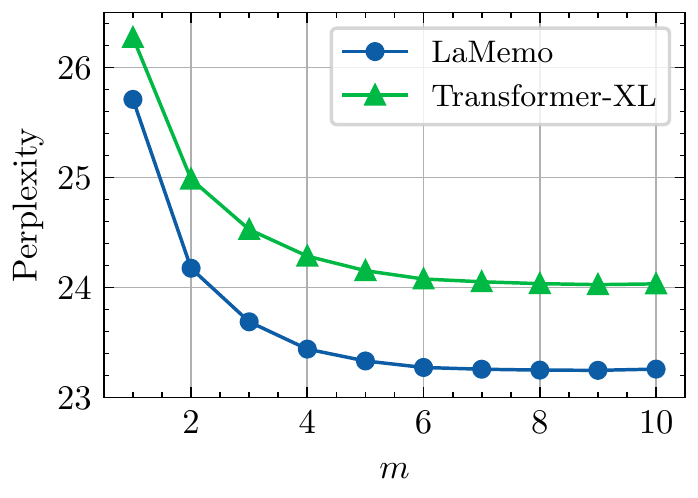}
    \caption{Test perplexity of LaMemo and Transformer-XL when extrapolating to longer contexts during inference, where $m$ is the ratio of the memory length to the target length.}
    \label{fig:extrap}
\end{figure}



In this section, we extrapolate the models to longer contexts during inference to study the effect of dynamic contextualization to the distant past.

We fix the length of the target sequence to $64$ and extrapolate the trained models to longer memory length $64\times m$ during inference, where $m = 1,\cdots,10$. We compare the perplexity of LaMemo and Transformer-XL trained on Wikitext-103 when augmented by a memory with different length. As shown in Figure \ref{fig:extrap}, LaMemo consistently achieves lower perplexity than Transformer-XL when extraploating to longer contexts, while the performance of both models saturate when $m$ is over 7. Additionally, we observe that the gap of perplexity between the two models increases when taking longer contexts into account. 
This demonstrates the effectiveness of dynamically refreshing the distant memory representations under the current context. 




\begin{figure}[t!]
    \centering
    \includegraphics[width=\columnwidth]{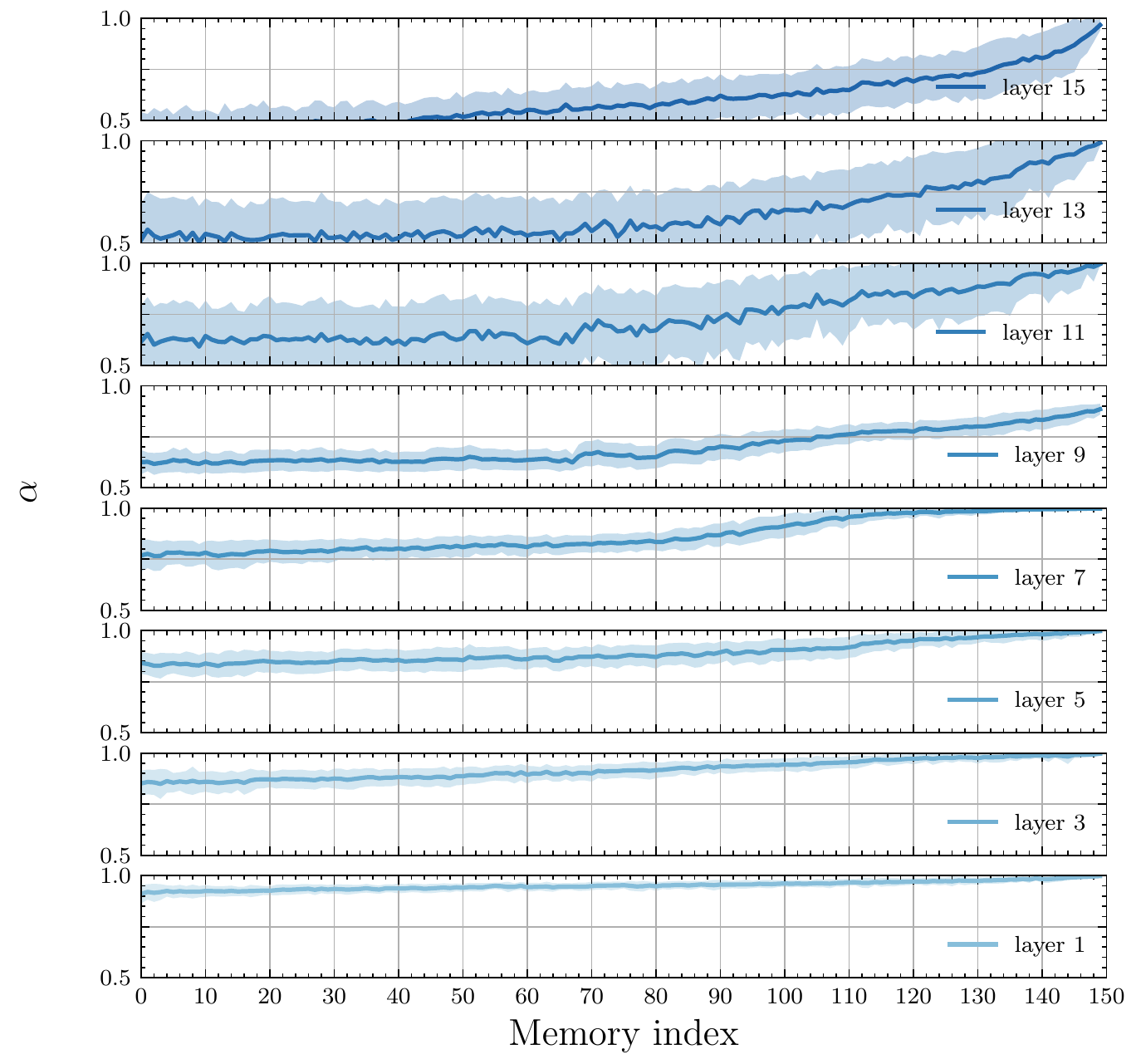}
    \caption{The memorizing coefficient $\alpha$ of different layers in a 16-layer model with a same memory and target length of 150. Smaller index means older memory.}
    \label{fig:alpha}
\end{figure}

\section{Attention Analysis}

In this section, we analyze the attention distribution of LaMemo to validate the effectiveness of utilizing bi-directional contexts with look-ahead attention.

We first visualize the memorizing coefficient $\alpha$ which stands for the portion of the past activations in the current memory representations. As show in Figure \ref{fig:alpha}, we plot $\alpha$ in different layers as a function of the memory index averaged on 100 text segments.\footnote{Due to the space limit, we only sample 8 layers from all the 16 layers.} We observe that in lower layers the memory mainly attends to the past ($\alpha \approx 1.0$). We conjecture that long-term bi-directionality is not necessary for low-level representations such as lexical features. In higher layers, the memory substantially utilizes the future contents to refresh the high-level representations, especially for the old memory state with a small memory index. 

Next, we visualize the attention weight distribution on the context tokens when predicting each target token 
in Figure \ref{fig:attn_utils}. For every token, we take the maximal attention weight in each interval of 5 tokens on its left and scale to a context length of 100. The result indicates that LaMemo learns better memory represetations by attending to the right-side tokens, which increases the memory utilization when predicting the target token.

\section{Case Study}

We present the generated texts of LaMemo and Transformer-XL trained on Wikitext-103 in Appendix \ref{app:gen_result}. 
Both models maintain a memory size of 512, and we seed them with the same context randomly sampled from the test set and generate 256 tokens using top-p sampling~\citep{neucleus_samp} with $p=0.95$.

\section{Related Work}

The Transformer~\citep{vaswani}, with its pair-wise modeling ability of the input, becomes prevailing for sequence modeling, especially long sequence processing tasks, such as long text generation~\citep{DBLP:conf/naacl/TanYAXH21,discodvt}, long document QA~\citep{longformer,etc}, language modeling~\citep{trm-xl,compressive}, video processing~\citep{DBLP:conf/cvpr/WuF0HKG19}, and etc. Specifically, language modeling~\citep{wikitext103} which requires processing documents with thousands of tokens has become a natural testbed for benchmarking this long-term processing ability. However, due to the quadratic time and space complexity of self-attention, scaling to inputs with thousands of tokens is computationally prohibitive.

One line of work investigated the linear-time attention mechanism to mitigate the scability issue of Transformer. Linformer~\citep{linformer} projects the inputs to lower dimension in length and approximates the full attention with a low-rank factorization. Linear Transformer~\citep{linear_transformer} regards the self-attention as a kernel function and uses a linear dot-product as a substitute. \citet{performer} and \citet{rfa} proposed to approximate the softmax more precisely with the expectation of the dot-product of random features. Although achieving substantial improvements on benchmarks designated for long inputs~\citep{lra}. These methods, however, focus on approximating the full attention with low-rank factorizations or kernel functions, which compromise the expressiveness and robustness of the original softmax attention, are reported to be inferior to the simple local attentions on real world language processing tasks~\citep{simple_local}.

Our work falls in another line, which augments the Transformer with a parametrized memory to store critical history information. Memory-augmented networks~\citep{neural_turing, mem_net_jason, mem_net_sukh} have been studied in the context of recurrent neural networks for a long time, but are mostly restricted to small and synthetic datasets. 
With the rapid development of Transformer, various works start to adapt memories to this architecture. 

\citet{trm-xl} first extended Transformer with a recurrence memory that caches hidden states computed in previous steps for the target tokens to attend to. \citet{compressive} further extended the context with an external memory that stores compressed hidden states at the temporal level. \citet{informer} used continuous space attention to attend over the old history and updated the memory with recent hidden states to enable unbounded memory capacity. \citet{wu2021memformer} proposed to use the encoder-decoder architecture to encode the memory states with previous text segments and pass this memory to future time steps. Instead of using a fixed-size attention span for different layers, \citet{adaptive_span} and \citet{adaptive_sparse} proposed to learn dynamic attention spans for different attention heads, which greatly reduced the computations. These works focused on enabling the Transformer to access contents in long distance, but did not consider to learn better memory representations by refreshing the old memory under the current context. Our work is orthogonal to learning adaptive attention spans and can be combined with this technique to reduce the complexity.  





\section{Conclusion}


We present LaMemo, a memory mechanism that allows the memory states to incrementally attend to the right-side tokens and interpolates with the old memory states on the left side, which enables the memory to interact with bi-directional contexts with a complexity linear in memory length. Experiments on three language modeling datasets demonstrate the superiority of LaMemo over baselines with various types of memory mechanisms. We also found that LaMemo increases the utilization of older memory states when predicting the target tokens, and yields a higher performance boost when extrapolating to longer memory length, which indicates the effectiveness of recontextualizing the memory under the current context.

\section*{Acknowledgments}

This work was supported by the National Science Foundation for Distinguished Young Scholars (with No. 62125604) and the NSFC projects (Key project with No. 61936010 and regular project with No. 61876096). This work was also supported by the Guoqiang Institute of Tsinghua University, with Grant No. 2019GQG1 and 2020GQG0005. This work was also sponsored by Tsinghua-Toyota Joint Research Fund. 

\bibliography{custom}
\bibliographystyle{acl_natbib}

\clearpage
\appendix
\label{sec:appendix}

\section{Derivation of Memory Interpolation}\label{app:derive}

We derive Eq. \ref{equ:mem-interp} into the form of standard self-attention in the following:
\begin{align}
    \boldsymbol{C}^{\leftrightarrow}_{\tau-1} &= \boldsymbol{\alpha}_{\tau-1}\text{sg}(\boldsymbol{C}_{\tau-1}^{\rightarrow}) + (1-\boldsymbol{\alpha}_{\tau-1})\boldsymbol{C}_{\tau-1}^{\leftarrow}. \nonumber
\end{align}
We consider the $i$-th row of $\boldsymbol{C}^{\leftrightarrow}_{\tau-1}$, denoted as $\boldsymbol{c}_i^{\leftrightarrow}$. We omit the stop-grad operation $\textrm{sg}(\cdot)$ and substitute $\alpha$ with the result from Eq. \ref{equ:alpha}:
\begin{align}
    \boldsymbol{c}_i^{\leftrightarrow} &= \alpha_i \boldsymbol{c}_i^{\rightarrow} + (1 - \alpha_i) \boldsymbol{c}_i^{\leftarrow} \nonumber \\
    &= \frac{s_{i}^{\rightarrow}}{s_i^{\rightarrow} + s_i^{\leftarrow}} \boldsymbol{c}_i^{\rightarrow} + \frac{s_{i}^{\leftarrow}}{s_i^{\rightarrow} + s_i^{\leftarrow}} \boldsymbol{c}_i^{\leftarrow} \nonumber,
\end{align}
where $s_i^{\rightarrow}$, $s_i^{\leftarrow}$ is the denominator of the softmax when computing $\boldsymbol{c}_i^{\rightarrow}$, $\boldsymbol{c}_i^{\leftarrow}$ respectively:
\begin{align}
    s_i^{\rightarrow} &= \sum_{j\le i}\exp\Big(\frac{\boldsymbol{q'}_i^\top \boldsymbol{k'}_j}{\sqrt{d}}\Big) = \sum_{j\le i} \text{sim}(\boldsymbol{q'}_i, \boldsymbol{k'}_j), \nonumber \\ 
    s_i^{\leftarrow} &= \sum_{j > i}\exp\Big(\frac{\boldsymbol{q}_i^\top \boldsymbol{k}_j}{\sqrt{d}}\Big) = \sum_{j > i} \text{sim}(\boldsymbol{q}_i, \boldsymbol{k}_j), \nonumber
\end{align}
where $(\boldsymbol{q}'_i, \boldsymbol{k}'_j)$ and $(\boldsymbol{q}_i, \boldsymbol{k}_j)$ are two sets of query-key vectors computed in the previous and this text segment respectively for the same position pair $(i,j)$ . Then we have:
\begin{align}
    &\boldsymbol{c}_i^{\leftrightarrow} = \frac{\sum_{j \le i} \text{sim}(\boldsymbol{q'}_i, \boldsymbol{k'}_j)}{\sum_{j \le i} \text{sim}(\boldsymbol{q'}_i, \boldsymbol{k'}_j) + \sum_{j > i} \text{sim}(\boldsymbol{q}_i, \boldsymbol{k}_j)} \boldsymbol{c}_i^{\rightarrow} \nonumber \\
    &+ \frac{\sum_{j > i} \text{sim}(\boldsymbol{q}_i, \boldsymbol{k}_j)}{\sum_{j \le i} \text{sim}(\boldsymbol{q'}_i, \boldsymbol{k'}_j) + \sum_{j > i} \text{sim}(\boldsymbol{q}_i, \boldsymbol{k}_j)} \boldsymbol{c}_i^{\leftarrow} \nonumber \\
    &=\frac{\sum_{j \le i} \text{sim}(\boldsymbol{q'}_i, \boldsymbol{k'}_j)\boldsymbol{v'}_j + \sum_{j > i} \text{sim}(\boldsymbol{q}_i, \boldsymbol{k}_j)\boldsymbol{v}_j}{\sum_{j \le i} \text{sim}(\boldsymbol{q'}_i, \boldsymbol{k'}_j) + \sum_{j > i} \text{sim}(\boldsymbol{q}_i, \boldsymbol{k}_j)} \nonumber \\
    &=\sum_{j} \beta_{j} \boldsymbol{\tilde{v}}_j , \nonumber
\end{align}
where $\sum_{j}\beta_j=1$. Finally, we derive $\boldsymbol{c}_i^{\leftrightarrow}$ as the weighted sum of the value vectors $\boldsymbol{\tilde{v}}_j$ from both the past ($j\le i$) and the future ($j > i$) of the position $i$.

\section{Unstability Analysis of the RPE in Transformer-XL}\label{app:unstability}

We conjecture that the unstability of Eq. \ref{equ:xl-rpe} stems from the terms involving the dot-product of $\boldsymbol{R}_{i-j}$ and another vector. So we start by considering the variance of $\boldsymbol{x}^\top \boldsymbol{R}_{i-j}$ where $\boldsymbol{x}\in\mathbb{R}^{d}$ is a random vector. Without loss of generality, we assume that $\boldsymbol{x}$ has zero mean and a variance of $\boldsymbol{\sigma}$:
\begin{align}
    \textrm{E}(x_k) &= 0, \ \forall k \in [1,\cdots, d] \nonumber \\
    \textrm{Var}(x_k) &= \sigma_{k,k}, \ \forall k \in [1,\cdots, d] \nonumber \\
    \textrm{Cov}(x_k, x_l) &= \sigma_{k,l}, \ \forall l \ne k \in [1,\cdots, d] \nonumber
\end{align}
Let $i-j=\Delta$. According to \citet{vaswani}, $\boldsymbol{R}_{\Delta}$ takes the following form:
\begin{align}
    \boldsymbol{R}_{\Delta}=&[\sin(\omega_1\Delta), \cos(\omega_1\Delta),\nonumber\\
    &\cdots,\sin(\omega_{d/2}\Delta), \cos(\omega_{d/2}\Delta)], \nonumber
\end{align}
where $w_k=10000^{-2k/d}$. Then the dot-product $\boldsymbol{x}^\top \boldsymbol{R}_{\Delta}$ can be derived into the linear combination of sine and cosine functions:
\begin{align}
    \boldsymbol{x}^\top \boldsymbol{R}_{\Delta} & = \sum_{k=1}^{d/2} x_{2k-1} \sin(\omega_k\Delta) + x_{2k} \cos(\omega_k\Delta), \nonumber
\end{align}
where we can easily derive that $\textrm{E}(\boldsymbol{x}^\top \boldsymbol{R}_{\Delta})=0$. According to the variance-expectation formula: $\textrm{Var}(x) = \textrm{E}[x^2] - \textrm{E}[x]^2$, we can simplify the variance $\textrm{Var}(\boldsymbol{x}^\top \boldsymbol{R}_{\Delta})$ in the following:
\begin{align}
    &\textrm{Var}(\boldsymbol{x}^\top \boldsymbol{R}_{\Delta})\nonumber\\
    &= \textrm{E}\Big[\big( \sum_{k=1}^{d/2} x_{2k-1} \sin(\omega_k\Delta) + x_{2k} \cos(\omega_k\Delta)\big)^2 \Big] \nonumber \\
    & = \sum_{k=1}^{d/2} \textrm{E}[x^2_{2k-1}]\sin^2(\omega_k\Delta) + \textrm{E}[x^2_{2k}]\cos^2(\omega_k\Delta) \nonumber \\
    &+ 2\sum_{k=1}^{d/2}\sum_{l=1,l\ne k}^{d/2} \textrm{E}[x_{2k-1}x_{2l}]\sin(\omega_k\Delta)\cos(\omega_l\Delta). \nonumber
\end{align}
We further simplify the above equation by assuming that all the elements have the same variance $\sigma_s$, and all pairs of distinct elements have the same covariance $\sigma_c$:
\begin{align}
    &\textrm{Var}(\boldsymbol{x}^\top \boldsymbol{R}_{\Delta}) =\sum_{k=1}^{d/2} \sigma_s [\sin^2(\omega_k\Delta) + \cos^2(\omega_k\Delta)] \nonumber \\
    &+ 2\sum_{k=1}^{d/2}\sum_{l=1,l\ne k}^{d/2} \sigma_c \sin(\omega_k\Delta)\cos(\omega_l\Delta) \nonumber \\
    &= \frac{d}{2}\sigma_s + 2\sigma_c g(\Delta), \nonumber
\end{align}
where $g(x) = \sum_{k=1}^{d/2}\sum_{l=1,l\ne k}^{d/2} \sin(\omega_kx)\cos(\omega_lx)$ is an odd function. 

\begin{figure}[t!]
    \centering
    \includegraphics[width=0.9\columnwidth]{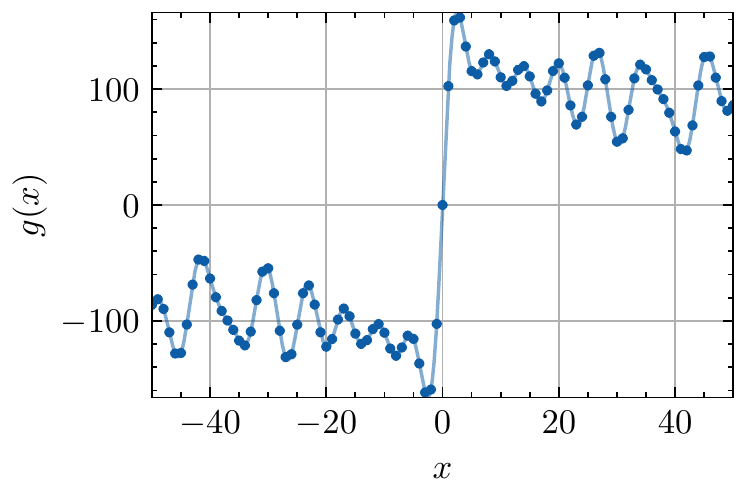}
    \caption{The plot of $g(x)$ when $d=64$. We see that $g(x)$ is symmetric with respect to the origin. The value of $g(x)$ when $x$ approaches zero from the left and right diverge greatly.}
    \label{fig:g(x)}
\end{figure}

We consider the value of $g(x)$ when $x\approx 0$. Since $\sin(\omega_k x) \approx \omega_k x$, $\cos(\omega_k x)\approx 1$, we have:
\begin{align}
    g(x)&\approx \sum_{k=1}^{d/2}\sum_{l=1}^{d/2} \omega_k x \nonumber \\
    &= \frac{d}{2}\sum_{k=1}^{d/2} w_k x \nonumber \\
    &= \frac{xd}{2}\sum_{k=1}^{d/2}\Big(\frac{1}{10000^{2/d}}\Big)^k \nonumber \\
    &\approx \frac{d}{2((10^8)^{1/d} - 1)}\cdot x = \gamma_d \cdot x. \nonumber
\end{align}
Since $a^x\approx1 + x \ln a$ when $x\approx 0$, we derive that $\gamma_d\approx \frac{d^2}{2 \ln 10^8}$ with the grow of $d$.  
This causes $g(x)$ to have a very steep slope near 0. Since $g(x)$ is an odd function, the value of $g(\Delta)$ and $g(-\Delta)$ will have a huge gap ($\Delta$ is a small positive value). To validate this, we plot the function of $g(x)$ when $d=64$ in Figure \ref{fig:g(x)}. 

Overall, the variance of $\boldsymbol{x}^\top \boldsymbol{R}_{\Delta}$ is composed of two terms, the first being $\sigma_s$ multiplied by a constant factor $d/2$, and the second being $\sigma_c$ multiplied by $g(\Delta)$. Note that $\sigma_s$ is strictly positive, while $\sigma_c$ does not have this restriction. Due the asymptotic behavior of $g(\Delta)$ near 0, i.e., $\mathcal{O}(d^2\Delta)$, we cannot find a proper $\sigma_c$ that makes $\textrm{Var}(\boldsymbol{x}^\top \boldsymbol{R}_{\Delta})$ bounded by $\mathcal{O}(d\sigma_s)$ for every $\Delta$ that takes its value from both the positive and negative integers.

\begin{figure}[t!]
    \centering
    \includegraphics[width=0.9\columnwidth]{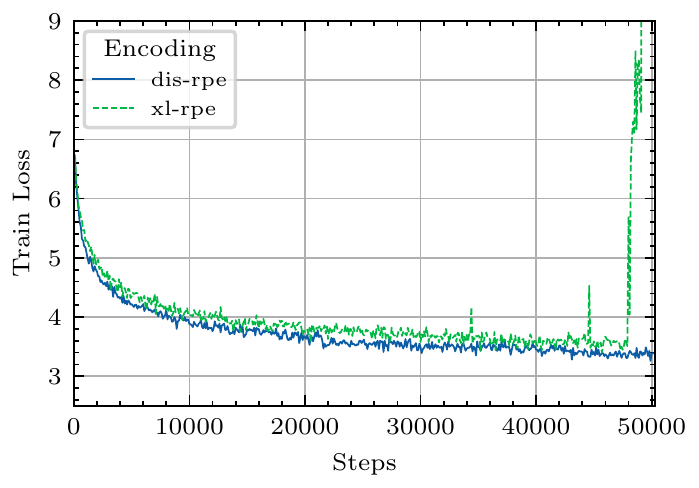}
    \caption{Comparison of the training dynamics using different encoding schemes: the disentangled RPE (\textbf{dis-rpe}) and the RPE of Transformer-XL (\textbf{xl-rpe}).}
    \label{fig:loss-comp}
\end{figure}

Finally, we plot the training curves of the two models using the RPE in Transformer-XL (xl-rpe) and our disentangled RPE (dis-rpe) in Figure \ref{fig:loss-comp} where we observed that the xl-rpe suffers from numerical unstability during training.


\section{Experimental Details}\label{app:experiment}

\subsection{Dataset Details}





\textbf{Wikitext-103} dataset is extracted from the set of verified Good and Featured articles on English Wikipedia. The dataset retains the original case, punctuation and numbers, and covers a broad range of domains, e.g., science, culture, bibliography, and etc. The dataset is available under the Creative Commons Attribution-ShareAlike (CC BY-SA) License.

\textbf{enwik8} dataset is the test set data of the Large Text Compression Benchmark which contains the first 100 million bytes of English Wikipedia dump on Mar. 3, 2006. All characters are encoded in UTF-8. This dataset is licensed under the CC BY-SA License.

\textbf{text8} dataset contains the first 100 million bytes of the clean text of Wikipedia that retains only regular articles and image captions. All the letters are converted into lower case, and only letters in the 27 character alphabet, namely letters \texttt{a-z} and nonconsecutive spaces, are preserved. This dataset is licensed under the CC BY-SA License.

The statistics of the three datasets is shown in Table \ref{tab:stats}.

\begin{table}[t!]
    \centering
    \small
    \begin{tabular}{l|c}
    \toprule[1pt]
    Dataset & train / dev / test \\
    \midrule[0.5pt]
    Wikitext-103 & 103,227,021 / 217,646 / 245,569 \\
    enwik8 & 88,982,818 / 4,945,742 / 4,943,417 \\
    text8 & 89,999,999 / 4,999,999 / 5,000,000 \\
    \bottomrule[1pt]
    \end{tabular}
    \caption{Statistics of the datasets used in the experiments. For Wikitext-103, we use the official split from \citet{wikitext103} and present the number of tokens in each split. For enwik8 and text8, we use the split from \citet{trm-xl} and report the number of characters for each split.}
    \label{tab:stats}
\end{table}

\subsection{Model Configurations}

We follow the base model configuration of \citet{trm-xl}. On Wikitext-103, we use the Transformer model with 16 layers, 10 attention heads with a head dimension of 41. The inner dimension size of the feedforward layer is 2100. We use a dropout rate of 0.1 and no attention dropout. To cope with the large vocabulary, we use the adaptive embeddings~\citep{adaptive_input}. We set the memory length to 150 and the target sequence length to 150 as well. On text8 and enwik8 datasets, we use the Transformer model with 12 layers, 8 attention heads with a head dimension of 64. The inner dimension size of the feedforward layer is 2048. We use a dropout rate of 0.1 and no attention dropout. We set the memory length to 512 and the target length to 512. Specifically, our LaMemo uses the disentangled relative positional encodings described in Sec. \ref{sec:dis-rpe}. The look-ahead attention shares the query, key and value projection matrices with those in the causal attention. 

\subsection{Training Settings}

We trained the models using Adam~\citep{adam} optimizer, with no warmup. We used a learning rate of $2.5\times 10^{-4}$ which decayed to 0 at the end of training with a cosine schedule. On Wikitext-103, we trained the model with 250K steps using a batch size of 64. On enwik8 and text8, we trained the model with 100K\footnote{We used a smaller number of training steps compared to \citet{trm-xl}, since it would take too long to train one model.} steps using a batch size of 40. We conducted our experiments on 2 Tesla V100. 

\subsection{Hyperparameters}



We present the hyperparameter search space in Table \ref{tab:hypram-search}. The number of hyperparameter search trials was 10. We adopted a manual search to select the hyperparameters, and the selection criterion was ppl/bpc on the dev set. We did not use early stopping during training.

\begin{table} [!htp]
\centering
\small
\setlength{\tabcolsep}{0.5mm}{
\begin{tabular}{cc}
\toprule[1pt]
Hyper-parameter & Search Space \\
\midrule[0.5pt]
Learning Rate & \textit{choice}[1e-4, \textbf{2.5e-4}, 5e-4] \\
Learning Rate Schedule & \textit{choise}[linear, \textbf{cosine}] \\
Warmup Steps & \textit{choice}[\textbf{0}, 1000, 2000] \\
Maximum Gradient Norm & \textit{choice}[\textbf{0.25}, 0.5, 1.0] \\
Epsilon (Sec. \ref{sec:mem_inter}) & \textit{choice}[1e-6, 1e-5, \textbf{1e-4}] \\
Optimizer & Adam \\
Epsilon (for Adam) & 1e-8 \\
Momentum (for Adam) & $\beta_1=0.9, \beta_2=0.999$ \\
\bottomrule[1pt]
\end{tabular}}
\caption{Hyperparameter search space. \textit{choice} indicates that the listed numbers will be chosen with the same probability. Best-found hyperparameters are in boldface.}
\label{tab:hypram-search}
\end{table}

\section{Generated Examples}\label{app:gen_result}

In this section, we present the examples generated by LaMemo and Transformer-XL trained on the Wikitext-103 dataset. Both models maintain a memory with a length of 512. We randomly select a piece of text from the test set as the context and allow both models to generate 256 tokens following the context. We use top-p sampling with $p=0.95$ and detokenize the context and the generated texts to facilitate reading. We present the exmples in Table \ref{tab:gen1} and \ref{tab:gen2}. We present our major findings below:
\begin{itemize}
    \item Both models are able to hallucinate imaginary contents fairly relevant to the limited contexts given as prompts. 
    \item Transformer-XL sometimes generates topic-irrelevant contents without further elaboration (marked by \underline{underline}), while LaMemo stays on topic more closely during the course of generation.
    \item Transformer-XL suffers more sever repetition issues (marked in \textbf{boldface}) than LaMemo both lexically and semantically.
\end{itemize}

\begin{table*}[t!]
    \centering
    \small
    \begin{tabularx}{2\columnwidth}{X}
    \toprule[1pt]
    \textbf{Context:}
    
    = Shackleton ( crater ) =

 \ \ Shackleton is an impact crater that lies at the south pole of the Moon. The peaks along the crater's rim are exposed to almost continual sunlight, while the interior is perpetually in shadow (a Crater of eternal darkness). The low-temperature interior of this crater functions as a cold trap that may capture and freeze volatiles shed during comet impacts on the Moon. Measurements by the Lunar Prospector spacecraft showed higher than normal amounts of hydrogen within the crater, which may indicate the presence of water ice. The crater is named after Antarctic explorer Ernest Shackleton.

 = = Description = =
 
 \ \ The rotational axis of the Moon lies within Shackleton, only a few kilometers from its center. The crater is 21 km in diameter and 4.2 km deep. From the Earth, it is viewed edge-on in a region of rough, cratered terrain. It is located within the South Pole-Aitken basin on a massif. The rim is slightly raised about the surrounding surface and it has an outer rampart that has been only lightly impacted. No significant craters intersect the rim, and it is sloped about 1.5 ° toward the direction 50 – 90 ° from the Earth. The age of the crater is about 3.6 billion years and it has been in the proximity of the south lunar pole for at least the last two billion years.

 \ \ Because the orbit of the Moon is tilted only 5 ° from the ecliptic, the interior of this crater lies in perpetual darkness. Estimates of the area in permanent shadow were obtained from Earth-based radar studies. Peaks along the rim of the crater are almost continually illuminated by sunlight , spending about 80 – 90 \% of each lunar orbit exposed to the Sun. Continuously illuminated mountains have been termed peaks of eternal light and have been predicted to exist since the 1900s.
 
 \ \ The shadowed portion of the crater was imaged with the Terrain Camera of the Japanese SELENE spacecraft using the illumination of sunlight reflected off the rim. The interior of the crater consists of a symmetrical 30 ° slope that leads down to a 6.6 km diameter floor. The handful of craters along the interior span no more than a few \\
 
    \midrule[0.5pt]
        
    \textbf{LaMemo Generation:}
    
    kilometres of plains. 
    
    \ \ A slight rotation is observed in the crater that now bears significant elements like snow, \textbf{ice}, and \textbf{water ice}. In other areas of the crater there are indications that the nature of the alpine ice layer provides some protection to the craters by increasing their brightness. The craters seem to be cut by an ice thickness of 10 cm (4 in). Images show their prominence, varying from a visible plateau to a multi-ring topography, indicating that they are present (a major component) in the craters. Their interior has a considerable surface area of dry ice. 
    
    \ \ As seen from Earth, the initial craters are marked by slow-moving metal-bearing mudstone-covered rock that has a surface of substantial amounts of coal, calcium, lead and sodium alloy, carbon monoxide, nitrogen oxides, and silicon dioxide. All are developed by processes such as heat transfer by the advancing radiation pressure. The least dense feature along the surface of the Moon is the large resurfacing event. There is no view that these rocks remain melting to about 90 °C (150 °F), but that their surface shine is dark and their outer layers are light, reflecting vertical oil. This late flank and early 1970s discovery was announced in 1982, as a substantial amount of ice. 
    
    \ \ A parallel lava \\
    
    \midrule[0.5pt]
    
    \textbf{{Transformer-XL} Generation:}
    
    kilometres. 
    
    \ \ During the hundreds of years after its formation, the crater now has significant seismic activity in the vicinity. Most craters are dome-shaped and are usually constructed of sedimentary materials that have been deposited during alpine ice times. The upper surface of the crater is covered with higher layers of rock that are cut by an extensive thickness of debris. The debris reaches the top of a mountain and comes in contact with visible ground planes. \underline{People are often} \underline{observed wearing headgear} of degrading materials such as \underline{clothing and boots, their shoes or hats}, or even working on the surface. \textbf{Below the crater}. As the \textbf{crater} faces the \textbf{crater} it has thick, thin pipes or scarps. A total of more than 200 caves have been excavated, down to some 40 m by 20 m. This exceeding the margin of the crater where it actually passes through is considered to be very high. Other geologic features by the advancing magnetic field have been reported from the crater. However, in 1992, scientists announced they would study this area again. 
    
    \ \ The crater was once a common feature of the Post Lunar System. Its medieval boundaries were not fixed in the orbital plane of Mercury. An individual crater had been called `` Discovery crater '' and one referred to as `` Bear crater '', although it is likely that an additional crater was called \\
    
    \bottomrule[1pt]
    
    \end{tabularx}
    \caption{Example 1 generated by LaMemo and Transformer-XL given a context prompt from the test set of Wikitext-103. Original Wikipedia page: \url{https://en.wikipedia.org/wiki/Shackleton_(crater)}.}
    \label{tab:gen1}
\end{table*}

\begin{table*}[t!]
    \centering
    \small
    \begin{tabularx}{2\columnwidth}{X}
    \toprule[1pt]
    
    \textbf{Context:}

     \ \ Nero was not expected to become Emperor because his maternal uncle, Caligula, had begun his reign at the age of 24 with enough time to produce his own heir. Nero 's mother, Agrippina, lost favour with Caligula and was exiled in 39 after her husband 's death. Caligula seized Nero 's inheritance and sent him to be brought up by his less wealthy aunt, Domitia <unk>, who was the mother of Valeria <unk>, Claudius 's third wife. Caligula, his wife <unk> and their infant daughter Julia Drusilla were murdered on 24 January 41. These events led Claudius, Caligula 's uncle, to become emperor. Claudius allowed Agrippina to return from exile.
     
     \ \ Claudius had married twice before marrying Valeria <unk>. His previous marriages produced three children including a son, Drusus, who died at a young age. He had two children with <unk> – Claudia Octavia (born 40) and Britannicus (born 41). <unk> was executed by Claudius in the year 48.
    
    \ \ In 49 AD , Claudius married a fourth time, to Nero 's mother Agrippina, despite her being his niece. To aid Claudius politically, young Nero was adopted in 50 and took the name Nero Claudius Caesar Drusus Germanicus (see adoption in Rome). Nero was older than his stepbrother Britannicus, and thus became heir to the throne. Nero was proclaimed an adult in 51 at the age of 14. He was appointed proconsul, entered and first addressed the Senate, made joint public appearances with Claudius, and was featured in coinage. In 53, he married his stepsister Claudia Octavia.
    
     = = Emperor (54 – 68 AD) = =
    
     = = = Early rule = = =
    
     \ \ Claudius died in 54 and Nero, taking the name Nero Claudius Caesar Augustus Germanicus, was established as Emperor. Though accounts vary, many ancient historians state Agrippina poisoned Claudius. According to Pliny the Elder, she used poison mushrooms. It is not known how much Nero knew or if he was even involved in the death of Claudius.
     
     \ \ Suetonius wrote `` ... for even if he was not the instigator of the emperor 's death , he was at least privy to it, as he openly admitted; for he used afterwards to laud mushrooms, the vehicle in which the poison was administered to Claudius, as ` the food of the gods , ' as the Greek proverb has it. At any rate, after Claudius' death he vented on him every kind of insult, in act and word, charging him now with folly and now with cruelty; for it was a favourite joke of his to say that Claudius had ceased ` to play the fool ' among mortals, lengthening the first syllable of the word <unk>, and he disregarded many of his decrees and acts as the work of a madman and a <unk>. Finally, he neglected to enclose the place where his body was burned except with a low and mean wall. ''
     
     \ \ According to Suetonius; Nero became Emperor at the age of 17 when the news of Claudius' death was made known , making him the youngest emperor at that time Although, what Suetonius may have meant is that he was \\
     
     \midrule[0.5pt]
     
     \textbf{LaMemo Generation:}
     
     killed as a result of the subsequent anti-Caligula propaganda. 
     
     \ \ At some time after 47 AD, Nero retired to his villa at Baiae. The combination of a curse that led to the death of Agrippa, and a duel between Claudius and his own son Drusus, may have encouraged the young Emperor to fire his own stela. 
     
     = = = = = Germanicus = = = = = 
     
     \ \ After becoming emperor, Nero received letters from Suetonius, which finally followed him on 28 May 44 AD with a letter to Claudius, completing the Emperor 's plot. The letters were printed in Latin. Claudius' plot was offered to Caligula by Agrippina the Elder, a woman from Kent, but she declined. Further letters from Gaius and Nero are known from the reigns of Tiberius and Caligul , though none of them mention Caligula. 
     
     \ \ Germanicus' conspiracy had developed into a protracted affair. His wife Antonia, his \textbf{sister} Agrippina 's \textbf{sister} and grandmother, had refused to accept him because she believed they were connected for communication, but the Via Flavia was left to her husband and Agrippina the Elder, whom she had defeated in her husband 's palace. Both had been sent through Gibraltar to compete for power. Gaius and Nero were outraged that Agrippina was supposed to be ready to support Caligula; Suetonius says that Nero insisted on making \\
     
     \midrule[0.5pt]
     
     \textbf{{Transformer-XL} Generation:}
     
     killed as a result of poison and was deemed a good \underline{comedian}. At now and then he had adopted Nero as his heir. 
     
     \ \ Although Nero 's rule was as close as he was to his father to his daughter Agrippina, he was not to physically exhibit any ill degree of higher kingship but to deny it. The Augustan History describes him as having been strongly willed, possessing an excellent hand and often claiming the title `` Augustus Caesar '', and often referred to him as Caesar 's `` paternal heir ''. The Augustan History, however, asserts that he was no longer in \underline{China}, and therefore he was raised as a Roman Hercules rather than a Roman citizen. \textbf{Claudius Claudius} was added as a junior emperor in 53 AD; he was crowned emperor in 61 AD. 
     
     = = = = As Augustus I = = = = 
     
     \ \ \textbf{Claudius} had developed a keen interest in \textbf{Claudius} in divination, a practice which later developed into his incarnation as a youth figure in Julius Caesar. Claudius ' grandfather, Leo I, ascended the throne in 23 AD and spent time in Rome, as did Claudius, who defeated Claudius in 42 AD. Claudius departed Rome after the death of Agrippa III in 65 AD. During the following years, Claudius was temporarily imprisoned in Rome, although possibly simply regulating the use of the captive dogs \\
     \bottomrule[1pt]
    \end{tabularx}
    \caption{Example 2 generated by LaMemo and Transformer-XL given a context prompt from the test set of Wikitext-103. Original Wikipedia page: \url{https://en.wikipedia.org/wiki/Nero}.}
    \label{tab:gen2}
\end{table*}

\end{document}